\documentclass[10pt,twocolumn,letterpaper]{article}
\usepackage{lineno}

\usepackage[pagenumbers]{cvpr} 

\definecolor{waymolgray}{HTML}{F0F0F0}  

\definecolor{cvprblue}{rgb}{0.21,0.49,0.74}
\usepackage[pagebackref,breaklinks,colorlinks,allcolors=cvprblue]{hyperref}
\usepackage{hyperref}       
\usepackage{url}            

\usepackage{booktabs}       
\usepackage{amsfonts}       
\usepackage{nicefrac}       
\usepackage{microtype}      
\usepackage{xcolor}         
\usepackage{graphicx}       
\usepackage{amsmath}
\usepackage{amssymb}
\usepackage{enumitem}
\usepackage{colortbl}       
\usepackage{pifont}         

\newcommand{\cmark}{\ding{51}} 
\newcommand{\xmark}{\ding{55}} 

\definecolor{waymogreen}{HTML}{00E89D}
\definecolor{waymolgreen}{HTML}{99F7D7} 
\definecolor{waymollgreen}{HTML}{CCFAEB} 
\definecolor{waymoblue}{HTML}{0077FF}
\definecolor{waymolblue}{HTML}{99B7FF}  
\definecolor{waymollblue}{HTML}{CCE4FF} 

\usepackage{soul} 

\def\paperID{14228} 
\def\confName{CVPR}
\def\confYear{2025}

\title{
FASIONAD : FAst and Slow FusION Thinking Systems for Human-Like Autonomous Driving with Adaptive Feedback}

\author{
    Kangan Qian$^{1*}$, Zhikun Ma$^{2*}$, Yangfan He$^{3,8*}$, Ziang Luo$^{1}$, Tianyu Shi$^{4}$, Tianze Zhu$^{1}$, \\
    Jiayin Li$^{5}$, Jianhui Wang$^{6}$, Ziyu Chen$^{7}$, Xiao He$^{7}$, Yining Shi$^{1}$, Zheng Fu$^{1}$, \\
    Xinyu Jiao$^{1}$, Kun Jiang$^{1}$, Diange Yang$^{1}$, Takafumi Matsumaru$^{2}$\\
    $^{1}$Tsinghua University, $^{2}$Waseda University, $^{3}$University of Minnesota - Twin Cities, \\
    $^{4}$University of Toronto, $^{5}$Xiamen University Malaysia,\\ 
    $^{6}$University of Electronic Science and Technology of China, $^{7}$AI2Robotics, \\ 
    $^{8}$Henan Runtai Digital Technology Group Co., Ltd. \\
    {\tt\small \{qka23, syn21, fu-z20, jiangkun, ydg, jiaoxinyu\}@mails.tsinghua.edu.cn,}\\
    {\tt\small zhikun@akane.waseda.jp,}
    {\tt\small he000577@umn.edu,}\\
    {\tt\small ty.shi@mail.utoronto.ca,}
    {\tt\small 20220991605023@std.uestc.edu.cn}\\
    \vspace{0.1cm}
}
\begin{document}
\maketitle
\renewcommand{\thefootnote}{}
\footnote{\textsuperscript{*}Equal contribution.}
\begin{abstract}

\noindent Ensuring safe, comfortable, and efficient navigation is fundamental to the development and reliability of autonomous driving systems. While end-to-end models trained on large datasets perform well in standard driving situations, they often struggle with rare, long-tail events. Recent advancements in large language models (LLMs) bring improved reasoning capabilities, yet their high computational demands complicate real-time decision-making and precise planning for autonomous vehicles. In this paper, we introduce \textbf{FASIONAD}, an innovative dual-system framework inspired by the cognitive model "Thinking, Fast and Slow." The fast system efficiently manages routine navigation tasks through rapid, data-driven path planning, while the slow system addresses complex reasoning and decision-making in unfamiliar or challenging scenarios. A dynamic switching mechanism, guided by score distribution and feedback, allows seamless transitions between fast and slow systems. Visual prompts from the fast system facilitate human-like reasoning in the slow system, which, in turn, supplies high-quality feedback to enhance the fast system’s decision-making. To evaluate our approach, we introduce a new benchmark derived from the nuScenes dataset, designed to distinguish between fast and slow scenarios. FASIONAD sets a new standard on this benchmark, pioneering a framework that differentiates fast and slow cognitive processes in autonomous driving. This dual-system approach offers a promising direction for creating more adaptive and human-like autonomous driving systems.

\end{abstract}   
\begin{figure*}[ht]
    \centering
    \includegraphics[width=\textwidth]{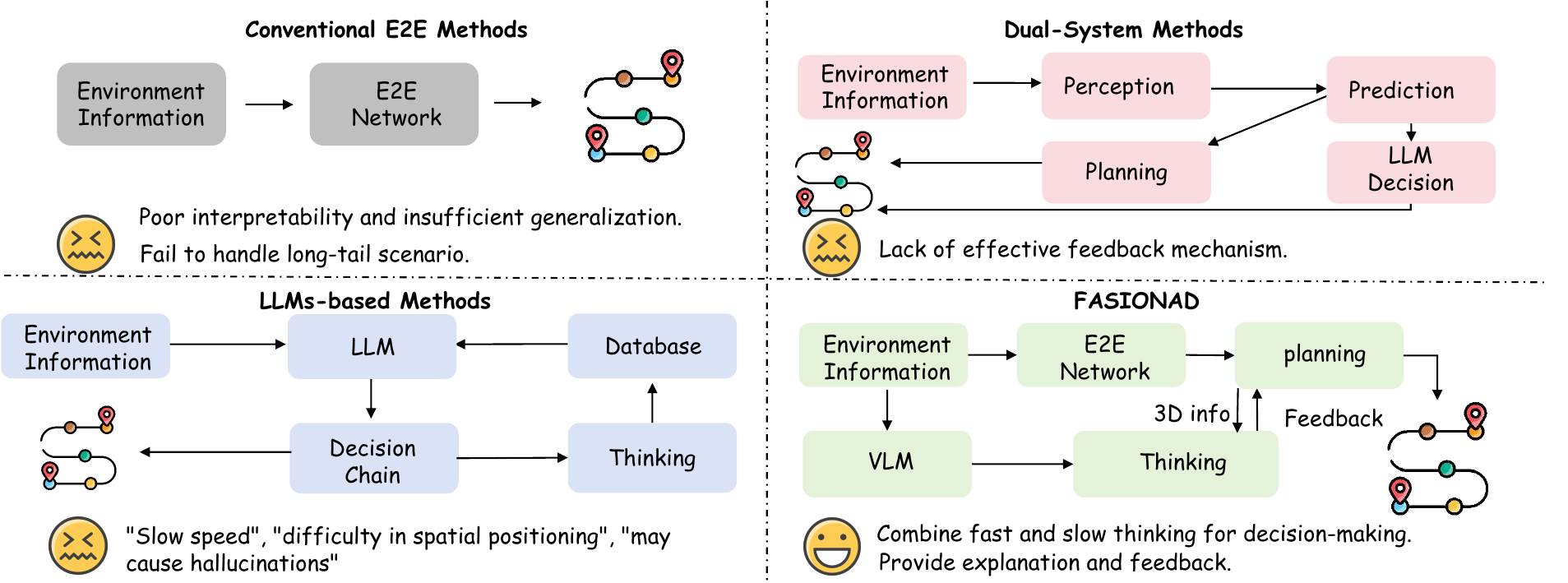}
    \caption{
   The motivation of our FASIONAD. Conventional E2E methods struggle with interpretability and generalization. LLMs-based methods face slow decision-making, spatial positioning issues, and potential hallucinations. The dual-system pipeline \cite{tian2024drivevlm} uses LLMs to fuse planning but lacks a safety feedback mechanism. We compares different motion planning methods for autonomous driving, showcasing our method’s ability to adaptive, context-aware decisions, offering better explanation and feedback. 
    }
    \label{fig:motivation}
    \vspace{-10pt}
\end{figure*}

\section{Introduction}

Autonomous driving has the potential to transform transportation by enhancing efficiency, reducing human workload, and minimizing accidents \cite{kiran2021deep}. Traditional autonomous driving systems typically adopt a modular design, with separate modules for perception, prediction, planning \cite{kiran2021deep}, and control. However, these systems struggle with adaptability in dynamic and complex environments, and face challenges in addressing long-tail problems and redundancy \cite{zhou2022dynamically, shi2019driving}, which limits their scalability and applicability.

To address these limitations, End-to-End (E2E) learning methods, such as Imitation Learning (IL) \cite{ALVLNN, hu2023planning, jiang2023vad, chen2024vadv2, PARA-drive} and Reinforcement Learning (RL) \cite{Reinforcement-Problem, e2e-challenge}, have been widely explored. However, IL methods are prone to covariate shift, leading to a lack of robustness in critical scenarios \cite{Imitation-Problem, liu2024curse}, even with improvements like Learning from Mistakes (LfM) \cite{arasteh2024validitylearningfailuresmitigating}. RL methods, while effective in simulations, face significant safety issues and encounter challenges in real-world applications, particularly due to difficulties in reward design and sim-to-real transfer \cite{sim2real}. Recent works such as DriveCoT \cite{wang2024drivecot} and DriveInsight \cite{driveinsight} aim to improve interpretability but are often time-consuming to generalize effectively across different scenarios.

With the recent advancements in Large Language Models (LLMs) and Vision-Language Models (VLMs), researchers have begun to explore their applications in autonomous driving, including manipulation tasks \cite{manipulation}, spatial grounding \cite{grounding}, and skill learning \cite{skill-learning}. However, despite these advancements \cite{sima2023drivelm, wang2023drivemlm, shao2024lmdrive, xu2024drivegpt4}, LLMs and VLMs still face challenges in spatial grounding and real-time decision-making \cite{zhang2024instruct}. Balancing safety and performance remains a critical issue \cite{wang2024safe}, which restricts their broader application in complex, real-world autonomous driving environments.


Inspired by the dual-process theory from psychology, particularly the concepts outlined in \textit{Thinking, Fast and Slow} \cite{booch2021thinking}, we introduce \textbf{FASIONAD}, an innovative \textbf{adaptive feedback} framework that seamlessly integrates fast and slow thinking methodologies. he framework employs \textbf{fast thinking} for routine driving tasks and leverages VLMs to handle the diverse complexities inherent in autonomous driving.

Within FASIONAD, we design a sophisticated switching mechanism that evaluates the complexity of each driving situation to determine whether to rely on the Fast Pathway or to engage the Slow Pathway. This mechanism is crucial for enabling adaptive, context-aware decision-making. By dynamically shifting between fast and slow pathways, FASIONAD ensures high responsiveness while maintaining safety in constantly changing environments.


We evaluate FASIONAD on challenging benchmarks, including \textit{nuScenes} and \textit{CARLA}. Experimental results demonstrate that FASIONAD outperforms state-of-the-art methods in both navigation success and safety. By enabling deeper situational understanding, intent analysis, and adaptive responses, FASIONAD marks a significant advancement in autonomous navigation. The main contributions of this work are as follows: \begin{itemize} \item We propose FASIONAD, an adaptive feedback framework for autonomous driving that combines fast and slow thinking to address the adaptability limitations of traditional systems in dynamic environments. \item We introduce a novel approach that simulates human decision-making, combining rapid strategy generation with VLM-based feedback evaluation to achieve optimal performance and flexibility in dynamic environments, addressing the limitations of pure end-to-end methods. \item Through extensive experiments on \textit{nuScenes} and \textit{CARLA}, we demonstrate that FASIONAD significantly enhances navigation success and safety, outperforming state-of-the-art methods by leveraging the principles of fast and slow thinking. \end{itemize}

\section{Related Work}
\setlength{\parskip}{0.5em} 
\subsection{Learning-based Planning}

Navigating dynamic and complex environments remains a central challenge in autonomous driving. Traditional methods often rely on modular pipelines that separate perception, planning, and control \cite{urmson2008autonomous, buehler2009darpa}, enabling targeted optimizations at each stage. However, these approaches can struggle with efficient information flow across modules, particularly in handling novel or unpredictable situations \cite{geiger2012we}. More recent advances have leaned towards end-to-end learning approaches, mapping raw sensor inputs directly to control commands \cite{bojarski2016end, codevilla2018end}. This trend is exemplified by GenAD \cite{zheng2024genad} and VAD \cite{jiang2023vad}, which integrate Bird's Eye View (BEV) representations with predictive models to navigate complex urban scenarios. Despite these advances, end-to-end methods often face challenges in interpretability and robustness when handling out-of-distribution samples \cite{chen2020learning, dosovitskiy2017carla}. Hybrid approaches that combine learning-based perception with classical rule-based planning have been explored to address these gaps. Furthermore, models like InterFuser and other transformer-based methods \cite{shao2023safety} have shown promise in leveraging multi-modal inputs for more refined decision-making. Recently, RL-GPT \cite{liu2024rlgpt} introduced a hierarchical framework that integrates reinforcement learning with task-specific code policies, achieving efficient decision-making in complex environments.

\subsection{Vision-Language Models for Reasoning}

VLMs have recently gained traction for enhancing perception in autonomous driving systems, providing a richer semantic understanding of complex scenes \cite{radford2021learning, alayrac2022flamingo}. These models align visual inputs with textual data, enabling more comprehensive scene interpretations, as seen in CLIP \cite{radford2021learning} and Flamingo \cite{alayrac2022flamingo}. Applications of VLMs in autonomous driving have focused on providing detailed scene descriptions and inferring intent \cite{li2023drivingclip, lin2023videollava}. For instance, Video-LLaVA \cite{lin2023videollava} and DrivingCLIP \cite{li2023drivingclip} enhance situational awareness through multi-turn dialogue capabilities and semantic scene understanding. The integration of VLMs into autonomous driving extends beyond perception, facilitating reasoning processes in dynamic environments. By interpreting interactions among traffic agents and extracting high-level semantic features, VLMs have been shown to improve decision-making in complex traffic scenarios \cite{fang2023video}. The FASIONAD framework advances these efforts by leveraging VLMs within its Thinking Module, using them not only for perception but also for reasoning during forward and backward planning. 

\begin{figure*}[ht]
    \centering
    \vspace{-10pt}
    \includegraphics[width=\textwidth]{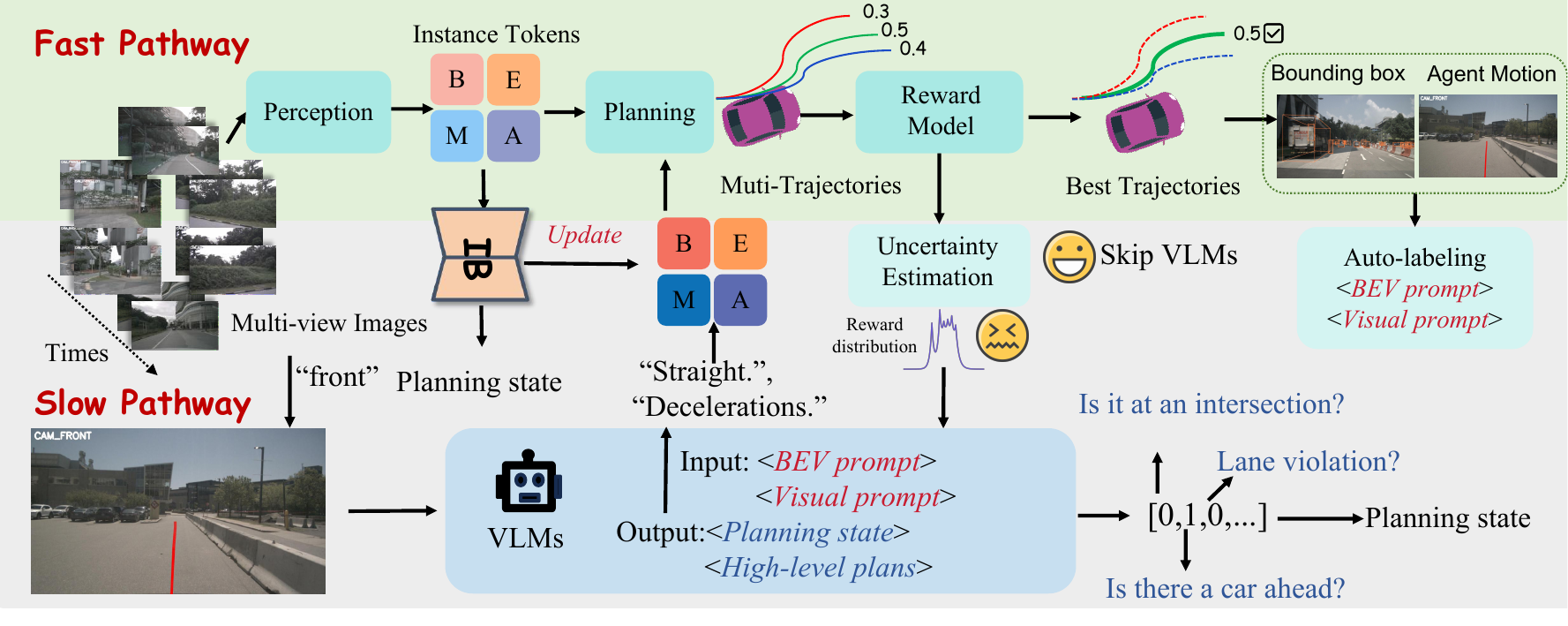}
    \vspace{-10pt}
    \caption{
    The framework operates through dual pathways: fast pathway and slow pathway. The fast pathway encodes image information into instance tokens(E, B, M, A relatively denotes ego tokens, BEV tokens, map tokens and agent tokens), generating multiple trajectories via a planning head. A reward model selects the optimal trajectory, while uncertainty estimation determines slow pathway activation. When engaged, the slow pathway utilizes VLM feedback, which is integrated both as augmented instance token queries and as scene-derived planning state vectors, enabling trajectory refinement through the planning head.
    }
    \label{fig:framework}
    \vspace{-10pt}
\end{figure*}

\section{Methodology}
\subsection{Overview}
As depicted in Fig.~\ref{fig:framework}, the FASIONAD framework employs a dual-pathway architecture: a Fast Pathway for rapid, real-time responses, and a Slow Pathway for comprehensive analysis and complex decision-making in uncertain or challenging driving scenarios. In the Fast Pathway, given a set of \( N \) multi-view images \( I_t = \{ I_{1t}, I_{2t}, \dots, I_{Nt} \} \) and high-level navigation commands \( \mathcal{C}_t \), the model generates a sequence of waypoints \( \mathcal{W}_t = \{ w_{1t}, w_{2t}, \dots, w_{Mt} \} \), where each waypoint \( w_{it} = (x_{it}, y_{it}) \) represents the predicted Bird’s Eye View (BEV) position of the ego vehicle at time \( t + i \). This pathway can be formulated as:
\begin{equation}
\text{FASIONAD (Fast Pathway):} \quad (I_t, \mathcal{C}_t) \rightarrow \mathcal{W}_t.
\end{equation}

In contrast, the Slow Pathway processes only the multi-view images \( I_t \) to generate a planning state \( \mathcal{P}_t \) and high-level meta-actions \( \mathcal{A}_t \), providing a more detailed assessment and strategic guidance for decision-making in complex scenarios. This pathway complements the Fast Pathway by enabling deeper analysis in uncertain or challenging conditions. The Slow Pathway is represented as:
\begin{equation}
    \text{FASIONAD (Slow Pathway):} \quad I_t \rightarrow (\mathcal{P}_t, \mathcal{A}_t).
\end{equation}

To coordinate Fast and Slow Pathways, we introduce uncertainty-based waypoints prediction and trajectory rewards. This mechanism optimizes responsiveness versus accuracy by dynamically activating either pathway based on environmental context and complexity, enabling both immediate reactions and thorough analysis when needed. Details of the Switch Mechanism are provided in Section \ref{fusion}.

\subsection{Fast Pathway}

\subsubsection{Waypoints Prediction and Reward Evaluation}
\textbf{Trajectory Generator.}
The trajectory generator outputs waypoint predictions \( \mathcal{W} = \{\mathbf{w}_t\}_{t=1}^{T} \), with each waypoint \( \mathbf{w}_t = (x_t, y_t) \) representing a spatial position in BEV coordinates. To capture interactions among traffic participants, we adopt a generative framework inspired by GenAD \cite{GenAD}, modeling trajectory prediction as a future trajectory generation problem.

\textbf{Reward Model.}  
The model generates \( N_C \times N_\mathcal{K} \) candidate trajectories \( \mathcal{T} = \{\mathbf{T}_i\}_{i=1}^{N_T} \), where each trajectory \( \mathbf{T}_i \in \mathbb{R}^{\text{bs} \times T_s \times 2} \) represents a sequence of waypoints over a time horizon \( T_s \). Here, \( N_C \) is the number of navigation commands, and \( N_\mathcal{K} \) represents the top-\( \mathcal{K} \) sampled multi-modal trajectories. Each trajectory \( \mathbf{T}_i \) is assigned a reward \( r_i \) by the reward model \( \mathcal{F}_{\text{Reward}} \), which integrates factors such as safety, comfort, efficiency, and economic considerations:
\begin{align}
    \mathcal{F}_{\text{Reward}} = &\ \alpha_{\text{safety}} C_{\text{safety}} 
    + \alpha_{\text{comfort}} C_{\text{comfort}} \notag \\
    &+ \alpha_{\text{efficiency}} C_{\text{efficiency}} 
    + \alpha_{\text{economic}} C_{\text{economic}}
\end{align}
where \( \alpha_{\text{safety}}, \alpha_{\text{comfort}}, \alpha_{\text{efficiency}}, \alpha_{\text{economic}} \) are weights determining the relative importance of each factor.

\textbf{Fast Pathway Loss Function.} We adopt the loss function design from \cite{jiang2023vad, GenAD}, which consists of a planning loss \(\mathcal{L}_{\text{plan}}\), an auxiliary 3D detection loss \(\mathcal{L}_{\text{det}}\), and a map segmentation loss \(\mathcal{L}_{\text{seg}}\). The total loss function is:
\begin{equation}
    \mathcal{L}_{\text{fast}} = \lambda_{\text{plan}}\mathcal{L}_{\text{plan}} + \lambda_{\text{det}} \mathcal{L}_{\text{det}} + \lambda_{\text{seg}}\mathcal{L}_{\text{seg}}
\end{equation}
where \(\lambda_{\text{plan}}, \lambda_{\text{det}}\), and \(\lambda_{\text{seg}}\) are weights balancing the auxiliary losses.

\subsection{Slow Pathway}
In complex scenarios, accurate interpretation of environmental factors is vital for safe decision-making. The Slow Pathway emulates human-like reasoning to infer context and predict future actions, similar to human drivers.This section discusses how VLMs can support such reasoning, with a focus on Question-Answering (QA) design in Section \ref{sec:QA} and data generation in Section \ref{sec:data-generation}.

\subsubsection{Planning-Oriented QA}
\label{sec:QA}
\begin{figure*}[htbp]
    \centering
    \vspace{-10pt}
    \includegraphics[width=\textwidth]{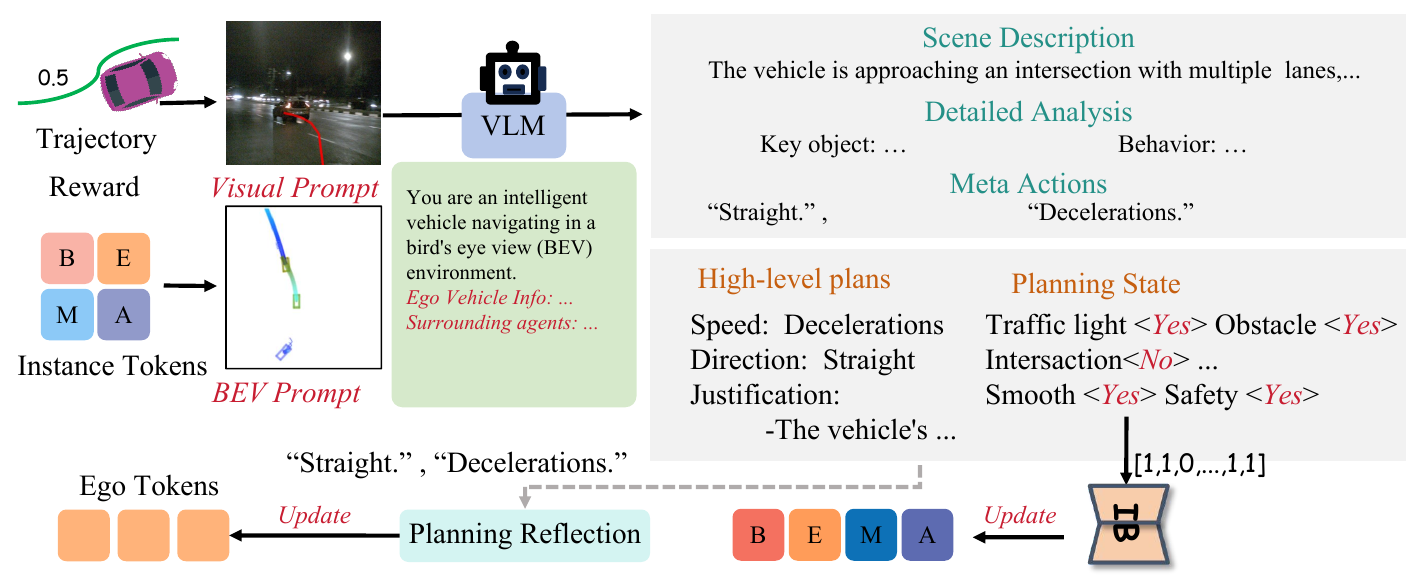}
    \vspace{-20pt}
    \caption{
    The Adaptive Feedback mechanism processes dual inputs: trajectory-generated images and BEV prompts derived from instance tokens, both feeding into a VLM. The VLM generates three distinct outputs: scene descriptions, analyses, and high-level plans, alongside planning state vectors that capture scene conditions. High-level plans are integrated into planning reflection, which modulates ego tokens, while planning state vectors are channeled through an information bottleneck to refine instance tokens.
    }
    \label{fig:vlm}
    \vspace{-10pt}
\end{figure*}

We propose a series of decision-oriented Question-Answering (QA) tasks to facilitate human-like reasoning in autonomous driving systems. Fig.~\ref{fig:vlm} illustrates the types of QA questions.

Our study addresses five key aspects essential for enhancing the robustness of autonomous driving systems by improving the system's understanding and replication of human-like driving behaviors:

\textbf{Scene Analysis.} This involves evaluating environmental factors such as weather conditions (e.g., sunny, rainy, snowy), time of day (morning, afternoon, evening, night), traffic density (light or heavy), and road conditions (wet, dry, icy). A thorough analysis of these factors enables the system to interpret the broader context, influencing critical decisions such as speed and maneuver selection.

\textbf{Traffic Sign Recognition.} This task focuses on recognizing and interpreting various traffic signs, including traffic lights, stop signs, yield signs, and speed limits. Accurate sign recognition is crucial for regulatory compliance and safety, forming a fundamental component of human-like driving behavior.

\textbf{Key Object Recognition and Behavior Analysis.} This involves identifying and analyzing key objects in the environment, such as vehicles, pedestrians, cyclists, and animals, and predicting their future behavior based on past movements. Accurate recognition and behavior prediction are vital for anticipating hazards and enabling proactive decision-making to avoid collisions.

\textbf{Planning State.} Planning-related states are represented as \( K \)-dimensional binary vectors that describe the current environmental context relevant to decision-making. This structured representation supports high-level planning by allowing the system to prioritize actions, optimize routing, and improve decision-making. Detailed state representations are provided in the supplementary materials.

\textbf{High-Level Planning and Justification.} This aspect involves formulating high-level plans for actions such as route selection, lane changes, and merging maneuvers, while considering long-term goals and constraints. By justifying these decisions, the system ensures that its actions are both safe and efficient, aligning with overarching driving objectives. This component is critical for replicating human-like decision-making in autonomous systems.

\subsubsection{Data Collection and Auto-labeling}
\label{sec:data-generation}
To generate these Question-Answering (QA) tasks, we utilize outputs from the Fast Pathway, including 3D object detection boxes and tracking trajectories, for automatic annotation. Additionally, we leverage Large Vision-Language Models (LVLMs), such as Qwen, to produce descriptive QAs that closely align with the observed scene and its elements. Inspired by the cognitive demands of driving decisions, we introduce two types of prompts to enhance QA generation: a visual prompt, which aids in interpreting visual cues and scene elements similarly to human perception, and a BEV prompt, which provides a top-down view of the environment to improve the understanding of spatial relationships and agent interactions.

To address the variability in VLMs outputs, which may contain extraneous or irrelevant information, we employ a regularization strategy inspired by few-shot learning in natural language processing (NLP). However, unlike general NLP applications, autonomous driving requires high reliability and consistency. Therefore, we refine the VLM outputs through a simplification process, ensuring that feedback to the Fast Pathway planner remains concise and effective, ultimately supporting the generation of new, accurate trajectories.

The slow pathway pipeline can be formulated as follows:
\begin{equation}
    \mathcal{P}_t, \mathcal{A}_t = \Phi[E(\mathcal{V}_t^{front}), E(\mathcal{B}_t)]
\end{equation}

\textbf{Visual Prompt.} In typical autonomous driving systems, waypoints generated by high-level planners are numerical outputs \cite{hu2023planning, jiang2023vad, chen2024vadv2}. However, VLMs are not inherently designed to process numerical data in this context. Human decision-making in complex driving scenarios relies more on intuitive reasoning and visual cues than on direct numerical computation. To bridge this gap, we integrate trajectory visual prompts into our slow-path planning. Specifically, we project the waypoints generated by the fast-path planner onto the front-view camera, creating a visual representation of the trajectory. This visual approximation of the planned path facilitates human-like reasoning processes, enabling more intuitive evaluation and modification of decisions, which leads to more reliable and effective high-level plans.

\textbf{BEV Prompt.} To further enhance the system's spatial understanding, particularly in 3D space and relative positioning, we introduce a BEV prompt. Based on the vehicle’s BEV coordinate system, this prompt provides a clear depiction of spatial relationships and dynamic interactions between the ego vehicle and surrounding agents, represented as \( bbox_{3D}^{Agent} \in \mathbb{R}^{N_{A} \times 7} \).

\textbf{Planning State and High-Level Plans.}
Planning states are represented as binary vectors \( \mathcal{P}_t = \{ 0, 1 \mid \mathcal{P}_t^j \in \mathbb{R}^{1 \times d_{\mathcal{P}}} \} \) and are determined through a \textit{Yes}/\textit{No} decision pipeline. Additionally, we propose a high-level plans encoder, denoted as \( E_{\mathcal{A}} \), which transforms high-level decisions from the VLM into meta-action features \( \mathcal{A}_t \). Since high-level plans can be decomposed into structured sets of meta-actions, the encoder \( E_{\mathcal{A}} \) performs a one-to-one mapping from these meta-actions to their corresponding meta-action features using a set of learnable embeddings \( e_{\mathcal{A}} \in \mathbb{R}^{N_{\mathcal{A}} \times d_{\mathcal{A}}} \), where \( N_{\mathcal{A}} \) represents the number of meta-actions. Finally, the planning state and meta-action features are input into the Fast Pathway to regenerate the trajectory, providing feedback that mimics human-like decision-making.



\textbf{Reward-Guided VLM Tuning.}
Traditional approaches with LLMs rely primarily on auto-regressive learning. In contrast, our approach combines auto-regressive learning with Maximum Likelihood Estimation (MLE) loss to tune the VLM. To improve prediction accuracy in complex scenarios, we introduce a reward-guided regression loss. Unlike InstructGPT \cite{mao2023gpt}, which depends on human feedback for reinforcement learning fine-tuning, our system utilizes automatically generated guidance. The objective is to replicate the planning state and high-level plans, which are directly accessible in our task setting. Thus, we define the ground truth as $[\mathcal{Y}_{\mathcal{P}_t}, \mathcal{Y}_{\mathcal{A}_t}]$. 

Since the GPT-based model typically applies supervision at the token level, whereas the entire sequence is meaningful for regression, we incorporate Proximal Policy Optimization (PPO) \cite{schulman2017ppo} with masking to apply supervision more effectively. The tuning loss, denoted as $\mathcal{L}_{\text{rvlm}}$, is computed as a reward within the policy gradient framework:
\begin{equation}
    \mathcal{L}_{\text{rvlm}} = \text{Reward}(\mathbf{s}^{1:T_i}) \cdot \Phi(\mathbf{s}^{T_i} | \mathbf{s}^{1:T_i-1})
\end{equation}
where $\mathbf{s}^{T_i}$ represents the predicted token at time step $T_i$, and $\text{Reward}(\mathbf{s}^{1:T_i})$ is the reward function for waypoint prediction in the Fast Pathway. The final training loss combines the standard language loss and the reward-guided loss:
\begin{equation}
\mathcal{L}_{\text{slow}} = \lambda_{\text{MLE}}\mathcal{L}_{\text{MLE}} + \lambda_{\text{rvlm}}\mathcal{L}_{\text{rvlm}}
\end{equation}

\subsection{FAst and Slow Fusion Autonomous Driving}\label{fusion}
\subsubsection{Uncertainty Estimation and Decision Mechanism}

To effectively navigate dynamic and unpredictable environments, estimating uncertainty in waypoint predictions is essential, as it allows the system to adapt its decision-making based on prediction reliability. To handle outliers and model uncertainty in waypoint predictions, we employ a Laplace distribution:
\begin{equation}
    p(\text{Reward}\mid \Theta) = \prod_{t=1}^{T} \frac{1}{2b} \exp\left( -\frac{\|\mathbf{r}_t - \hat{\mathbf{r}}_t\|_1}{b} \right)
\end{equation}

where \( \hat{\mathbf{r}}_t \) denotes the predicted reward at time \( t \), \( b \) is the scale parameter, and \( \Theta \) represents the model parameters. This distribution’s heavy tails make it robust to outliers, which is advantageous in dynamic driving environments.


 The Laplace distribution's heavy tails and sharp peak make it robust to outliers and effective for uncertainty estimation in dynamic driving environments. Based on the reward ( R ) and estimated uncertainty, the system chooses between the Fast Pathway for immediate navigation (when ( R ) exceeds a threshold and uncertainty is low) or the Slow Pathway for detailed analysis.

\vspace{-15pt}
\subsubsection{Feedback with Information Bottleneck}

Driving environments often contain significant irrelevant or noisy information that does not contribute to planning. To address this, we apply the information bottleneck principle \cite{planKD} to distill only the information relevant to decision-making. This approach ensures that the model prioritizes critical features for navigation, effectively minimizing the influence of extraneous data.

To align instance-aware features \( z \) with \( y_t \), we employ an MLP \( f_{\text{MLP}} \) that maps \( z \) to a one-dimensional vector \( y_i \). The knowledge distillation process minimizes the following objective:
\begin{equation}
\mathcal{L}_{\text{KD}} = \sum \log q_d(y_t | y_i) - \beta \, \text{KL}\left(q_e(y_i | z_{\text{current}}) \| p(z)\right)
\end{equation}
where \( q_d(y_t | y_i) \) is the probability distribution over the VLM-derived vector \( y_t \) given \( y_i \), and \( q_e(y_i | z_{\text{current}}) \) encodes instance-aware features from the current state. Here, \( p(z) \) is a prior distribution on \( z \), and \( \beta \) is a regularization parameter.

\subsubsection{Feedback Fusion Mechanism}

\begin{figure*}[ht]
    \centering
    \includegraphics[width=\linewidth]{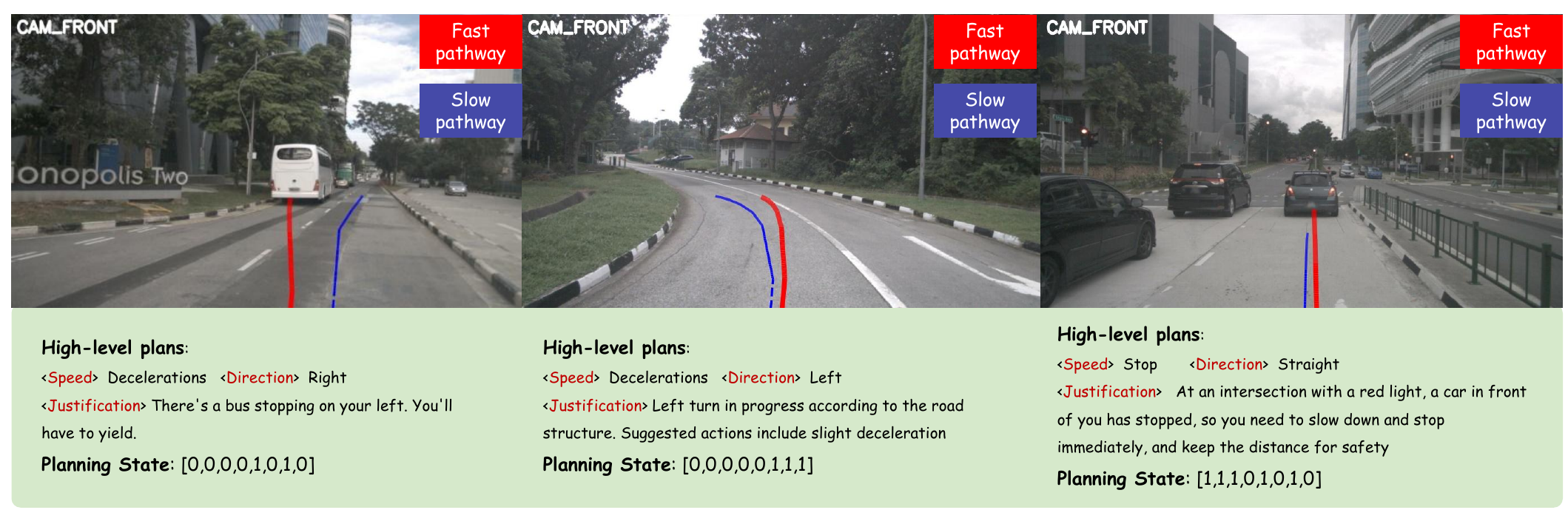}
    \caption{Example scenarios demonstrating FASIONAD's adaptive feedback framework in various driving environments. Each scene shows different navigation challenges, including obstacles, lane adjustments, and turns. The proposed system provides suggested driving operations and ensures safe, smooth trajectories with minimal abrupt maneuvers, enhancing navigation performance and safety in complex situations.}
    \label{fig:adaptivefeedbackl}
    \vspace{-10pt}
\end{figure*}





The Slow Pathway, activated by reward signals and uncertainty, enables selective deep analysis of essential VLM-derived features. Integration occurs through cross-attention between learnable embeddings \( e_{\mathcal{A}} \in \mathbb{R}^{N_A \times d_A} \) and the ego token \( e_{\text{ego}} \in \mathbb{R}^{d_A} \), where \( e_{\text{ego}} \) queries \( e_{\mathcal{A}} \) as key-value pairs. This captures contextual dependencies, and the resulting fused state feeds into the Fast Pathway for trajectory planning, mimicking human decision-making in complex driving scenarios.



\section{Experiments}
In this section, we evaluate the performance of our proposed \textbf{FASIONAD} framework on standard autonomous driving benchmarks. Specifically, we assess the system’s ability to navigate complex, dynamic environments while adhering to strict safety constraints. We focus on two widely used datasets: \textit{CARLA}~\cite{carla_benchmark} and \textit{nuScenes}~\cite{nuscenes_benchmark}, which provide realistic driving scenarios and varying environmental conditions.

\subsection{Experimental Setup}

Our evaluation for FASIONAD encompasses both open-loop and closed-loop performance metrics. For open-loop assessment, we utilize the nuScenes dataset, which provides comprehensive annotated data from urban driving scenarios. This evaluation focuses on measuring the policy's similarity to expert demonstrations through L2 distance and collision rate metrics. We prioritize these open-loop measurements in our ablation studies due to their computational efficiency and consistency in results.
The closed-loop evaluation employs the CARLA Closed-loop Town05 Short Benchmark, which features challenging scenarios including narrow streets, dense traffic, and frequent intersections. The primary performance indicators are the Driving Score (DS)—calculated as the product of Route Completion (RC) and Infraction Score—and Route Completion itself. To ensure fair comparison with existing methods, we implement a rule-based wrapper around the learning-based policy, following standard practice in benchmark evaluations. This wrapper helps minimize infractions during testing.
\begin{table*}[htbp]
\centering
\caption{Open-loop planning performance. Comparisons with state-of-the-art methods in motion planning performance on the nuScenes validation dataset. * denotes using ego status features as input. † represents that the metrics are computed with an average of all the predicted frames. † denotes FPS measured in the same environment on our machine with a single RTX 3090 GPU. We demonstrate superior end-to-end planning effectiveness and maintain competitive inference speed on the nuScenes validation dataset.}
\label{table:nuscenes_comparison}
\small
\begin{tabular}{lcccccccccc}
\toprule
\textbf{Method} & \textbf{Input} & \multicolumn{4}{c}{\textbf{L2 (m) $\downarrow$}} & \multicolumn{4}{c}{\textbf{Collision Rate (\%) $\downarrow$}} & \textbf{FPS} \\
\cmidrule(lr){3-6} \cmidrule(lr){7-10}
& & \textit{1s} & \textit{2s} & \textit{3s} & \textit{Avg} & \textit{1s} & \textit{2s} & \textit{3s} & \textit{Avg} & \\
\midrule
IL\cite{ratliff2006maximum} & LiDAR & 0.44 & 1.15 & 2.47 & 1.35 & 0.08 & 0.27 & 1.95 & 0.77 & - \\
NMP\cite{NMP} & LiDAR & 0.53 & 1.25 & 2.67 & 1.48 & 0.04 & 0.12 & 0.87 & 0.34 & - \\
FF\cite{FF} & LiDAR & 0.55 & 1.20 & 2.54 & 1.43 & 0.06 & 0.17 & 1.07 & 0.43 & - \\
EO\cite{EO} & LiDAR & 0.67 & 1.36 & 2.78 & 1.60 & 0.04 & 0.09 & 0.88 & 0.33 & - \\
ST-P3 ~\cite{hu2022stp3} & Camera & 1.33 & 2.11 & 2.90 & 2.11 & 0.23 & 0.62 & 1.27 & 0.71 & 1.6 \\
UniAD ~\cite{hu2023planning} & Camera & 0.48 & 0.96 & 1.65 & 1.03 & 0.05 & 0.17 & 0.71 & 0.31 & 1.8 \\
OccNet \cite{OCCNet} & Camera & 1.29 & 2.13 & 2.99 & 2.14 & 0.21 & 0.59 & 1.37 & 0.72 & 2.6 \\
VAD-Tiny ~\cite{jiang2023vad} & Camera & 0.60 & 1.23 & 2.06 & 1.30 & 0.31 & 0.53 & 1.33 & 0.72 & 6.9† \\
VAD-Base ~\cite{jiang2023vad} & Camera & 0.54 & 1.15 & 1.98 & 1.22 & 0.04 & 0.39 & 1.17 & 0.53 & 3.6† \\
GenAD \cite{GenAD} & Camera & 0.36 & 0.83 & 1.55 & 0.91 & 0.06 & 0.23 & 1.00 & 0.43 & 6.7† \\
\midrule\midrule
DriveVLM* \cite{tian2024drivevlm} & Camera & 0.15 & 0.29 & 0.48 & 0.31 & 0.05 & 0.08 & 0.17 & 0.10 & 6.7† \\
\cellcolor{waymolgray}\textbf{FASIONAD *(Ours)} & \cellcolor{waymolgray}Camera & \cellcolor{waymolgray}\textcolor{waymoblue}{\textbf{0.13}} & \cellcolor{waymolgray}\textcolor{waymoblue}{\textbf{0.26}} & \cellcolor{waymolgray}\textcolor{waymoblue}{\textbf{0.45}} & \cellcolor{waymolgray}\textcolor{waymoblue}{\textbf{0.28}} & \cellcolor{waymolgray}\textcolor{waymoblue}{\textbf{0.05}} & \cellcolor{waymolgray}\textcolor{waymoblue}{\textbf{0.08}} & \cellcolor{waymolgray}\textcolor{waymoblue}{\textbf{0.15}} & \cellcolor{waymolgray}\textcolor{waymoblue}{\textbf{0.09}} & \cellcolor{waymolgray}\textcolor{waymoblue}{\textbf{6.9}}† \\
\midrule\midrule
\textbf{Agent-Driver\cite{mao2024languageagent}} & Camera &  {\textbf{0.22}} &  {\textbf{0.65}} &  {\textbf{1.34}} &  {\textbf{0.74}} &  {\textbf{0.02}} &  {\textbf{0.13}} &  {\textbf{0.48}} &  {\textbf{0.21}} & None† \\
\cellcolor{waymolgray}\textbf{FASIONAD (Ours)} & \cellcolor{waymolgray}Camera & \cellcolor{waymolgray}\textcolor{waymoblue}{\textbf{0.19}} & \cellcolor{waymolgray}\textcolor{waymoblue}{\textbf{0.62}} & \cellcolor{waymolgray}\textcolor{waymoblue}{\textbf{1.25}} & \cellcolor{waymolgray}\textcolor{waymoblue}{\textbf{0.69}} & \cellcolor{waymolgray}\textcolor{waymoblue}{\textbf{0.02}} & \cellcolor{waymolgray}\textcolor{waymoblue}{\textbf{0.09}} & \cellcolor{waymolgray}\textcolor{waymoblue}{\textbf{0.44}} & \cellcolor{waymolgray}\textcolor{waymoblue}{\textbf{0.18}} & \cellcolor{waymolgray}\textcolor{waymoblue}{\textbf{6.9}}† \\
\bottomrule
\end{tabular}
\end{table*}

\subsection{Results and Analysis}
\definecolor{bestgreen}{rgb}{0.0, 0.6, 0.0}
\begin{table}[ht]
\centering
\caption{Closed-loop evaluation on Town05 Short benchmark. Enhanced Driving Performance and Infraction Analysis in challenging environments.}
\vspace{-5pt}
\footnotesize  
\setlength{\tabcolsep}{4pt} 
\label{tab:driving_methods_comparison}
\begin{tabular}{lccc}
\hline
\textbf{Methods} & \textbf{Modality} & \textbf{DS (\%) $\uparrow$} & \textbf{RC (\%) $\uparrow$} \\ \hline
CILRS ~\cite{codevilla2019exploring} & C &7.47 & 13.40 \\ 
LBC  ~\cite{cui2021lookout} & C &30.97 & 55.01 \\ 
Transfuser ~\cite{chitta2022transfuser} & C+L &54.52 & 78.41 \\ 
ST-P3 ~\cite{hu2022stp3} & C &55.14 & 86.74 \\ 
VAD ~\cite{jiang2023vad} & C &64.29 & 87.26 \\ 
Agent-Driver ~\cite{mao2024languageagent}& C &64.31 & 87.31 \\ 
\textbf{FASIONAD (Ours)} & C & \textcolor{waymoblue}{\textbf{64.83}} & \textcolor{waymoblue}{\textbf{89.04}} \\ \hline
\end{tabular}
\vspace{-20pt}
\end{table}

We compare FASIONAD’s performance against several baseline models, including ST-P3~\cite{hu2022stp3}, an end-to-end vision-based framework integrating spatial-temporal learning for improved real-time driving accuracy and safety; VAD~\cite{jiang2023vad}, a vectorized scene representation framework enabling efficient perception, prediction, and planning for faster decision-making without deep reinforcement learning; UniAD~\cite{hu2023planning}, a unified framework that prioritizes perception, prediction, and planning tasks, demonstrating superior performance on the nuScenes benchmark; CILRS~\cite{codevilla2019exploring}, a behavior cloning framework that addresses dataset bias and generalization issues in unseen environments; and Transfuser~\cite{chitta2022transfuser}, which uses a transformer-based fusion mechanism to integrate image and LiDAR data for enhanced perception in dense, dynamic scenarios. Additionally, we assess forward-planning-only systems, which generate plans without backward safety evaluations. We also conduct ablation studies to evaluate FASIONAD's performance without the slow strategy to measure the impact of safety evaluations and without LLM-based intent inference by substituting a rule-based system to quantify the LLM's contribution to understanding high-level driving intent.
\subsubsection{Baseline Driving Performance Comparison}
\textbf{Open-loop evaluation:} The nuScenes dataset was used to evaluate open-loop autonomous driving systems, covering diverse real-world scenarios like varying weather, traffic density, and urban complexity. We compared multiple methods based on L2 error and collision rate in Table \ref{table:nuscenes_comparison}, revealing their strengths and limitations under different conditions and highlighting their potential real-world applicability.
In the open-loop planning evaluation on the nuScenes validation dataset, FASIONAD achieved the best overall performance compared to state-of-the-art methods. It recorded the lowest average L2 distance of 0.80 meters and the lowest average collision rate of 0.32\%, demonstrating superior motion planning accuracy and safety.

FASIONAD outperformed strong baselines such as GenAD, which had an average L2 distance of 0.91 meters and a collision rate of 0.43\%, and VAD-Base, with an L2 distance of 1.22 meters and a collision rate of 0.53\%. FASIONAD also maintained competitive inference speed with 6.5 FPS, making it highly effective for real-time applications.

\textbf{Closed-loop evaluation:} We used the CARLA Closed-loop Town05 Short Benchmark to assess driving in tight, complex environments with short routes, multiple intersections, and dense traffic. The evaluation focuses on Driving Score (DS) and Route Completion (RC) to measure performance in fast-paced, dynamic scenarios.
In the closed-loop evaluation of autonomous driving methods on the Town05 Short benchmark, FASIONAD demonstrated the best overall performance. Known for its challenging road conditions, this benchmark tested the planning and driving capabilities of each method.

FASIONAD achieved a Driving Score (DS) of 64.83\% and a Rate of Completion (RC) of 89.04\%, outperforming all other methods. Notably, it surpassed VAD, which recorded a DS of 64.29\% and an RC of 87.26\%, and ST-P3, with a DS of 55.14\% and an RC of 86.74\%. These results highlight FASIONAD's superior ability to handle complex driving environments and complete routes effectively.


\subsubsection{Ablation Study}


\textbf{Information Bottleneck and High-Level Action Performance.} Our ablation study demonstrates the complementary benefits of the Information Bottleneck (IB) and High-Level Action (HA) components(Table~\ref{tab:info_bottleneck_high_level_action}). The full model incorporating both components achieved the best performance (L2: 0.69m, collision rate: 0.18\%). Using either component alone led to decreased performance - IB-only (L2: 0.74m, collision rate: 0.21\%) and HA-only (L2: 0.77m, collision rate: 0.19\%) - highlighting their synergistic relationship in improving prediction accuracy through effective information filtering and high-level planning.

\begin{table}[htbp]
\caption{Open-loop Validation of Information Bottleneck~(IB) and High-Level Action(HA). The models listed here are implemented without ego-state for purely evaluation of components.}
\footnotesize  
\setlength{\tabcolsep}{4pt} 
\centering
\small 
\begin{tabular}{cc|cccc|cccc}
\toprule
\multicolumn{2}{c|}{\textbf{ Setting}} & \multicolumn{4}{c|}{\textbf{L2 (m) $\downarrow$}} & \multicolumn{4}{c}{\textbf{Collision Rate (\%) $\downarrow$}} \\
\cmidrule(lr){1-2} \cmidrule(lr){3-6} \cmidrule(lr){7-10}
\textbf{IB} & \textbf{HA} & \textbf{1s} & \textbf{2s} & \textbf{3s} & \cellcolor{waymolgray}\textbf{Avg.} & \textbf{1s} & \textbf{2s} & \textbf{3s} & \cellcolor{waymolgray}\textbf{Avg.} \\
\midrule
\cellcolor{waymolgray}\cmark & \xmark & 0.23 & 0.66 & 1.34 & \cellcolor{waymolgray}0.74 & 0.03 & 0.12 & 0.47 & \cellcolor{waymolgray}0.21 \\
\xmark & \cellcolor{waymolgray}\cmark & 0.24 & 0.68 & 1.37 & \cellcolor{waymolgray}0.77 & \textbf{0.02} & 0.10 & 0.45 & \cellcolor{waymolgray}0.19 \\
\cellcolor{waymolgray}\cmark & \cellcolor{waymolgray}\cmark & \textbf{0.19} & \textbf{0.62} & \textbf{1.25} & \cellcolor{waymolgray}\textbf{0.69} & \textbf{0.02} & \textbf{0.09} & \textbf{0.44} & \cellcolor{waymolgray}\textbf{0.18} \\

\bottomrule
\end{tabular}
\label{tab:info_bottleneck_high_level_action}
\end{table}


\textbf{VLM Prompt Strategy Performance.} Our ablation study on VLM prompt strategies revealed the significant impact of prompt design (Table~\ref{tab:vlm_io_effect}). The Full.P configuration, featuring comprehensive prompt instructions, achieved the best results with an L2 distance of 0.69 meters and 0.18\% collision rate. Performance gradually declined with simpler prompting approaches: Visual.P (0.74m, 0.20\%), BEV.P (0.79m, 0.24\%), and Simple.P (0.80m, 0.32\%). These results demonstrate that detailed, well-structured prompts are crucial for maximizing VLM's predictive capabilities.

\begin{table}[htbp]
\caption{Validation of VLM Prompt Strategies. Impact of different prompting approaches on L2 error and collision rate. The models listed here are implemented without ego-state for purely evaluation of components.}
\footnotesize 
\setlength{\tabcolsep}{4pt} 
\centering
\small 
\begin{tabular}{c|cccc|cccc}
\toprule
\textbf{Setting} & \multicolumn{4}{c|}{\textbf{L2 (m) $\downarrow$}} & \multicolumn{4}{c}{\textbf{Collision Rate (\%) $\downarrow$}} \\
\cmidrule(lr){2-5} \cmidrule(lr){6-9}
& \textbf{1s} & \textbf{2s} & \textbf{3s} & \cellcolor{waymolgray}\textbf{Avg.} & \textbf{1s} & \textbf{2s} & \textbf{3s} & \cellcolor{waymolgray}\textbf{Avg.} \\
\midrule
Simple.P& 0.31 & 0.71 & 1.38 & \cellcolor{waymolgray}0.80 & 0.05 & 0.16 & 0.74 & \cellcolor{waymolgray}0.32 \\
BEV.P  & 0.29 & 0.70 & 1.36 & \cellcolor{waymolgray}0.79 & 0.04 & 0.14 & 0.65 & \cellcolor{waymolgray}0.24 \\
Visual.P  & 0.24 & 0.67 & 1.30 & \cellcolor{waymolgray}0.74 & \textbf{0.02} & 0.11 & 0.48 & \cellcolor{waymolgray}0.20 \\
Full.P & \textbf{0.19} & \textbf{0.62} & \textbf{1.25} & \cellcolor{waymolgray}\textbf{0.69} & \textbf{0.02} & \textbf{0.09} & \textbf{0.44} & \cellcolor{waymolgray}\textbf{0.18} \\
\bottomrule
\end{tabular}
\label{tab:vlm_io_effect}
\end{table}

\section{Conclusion and Future Work}

In summary, \textbf{FASIONAD} establishes a robust foundation for autonomous navigation, effectively integrating fast-response capabilities and safety-focused assessments within a modular, scalable framework. Achieving a \textbf{10-15\% reduction in collision metrics} compared to state-of-the-art baselines and reaching a \textbf{Driving Score (DS) of 64.83\%} and \textbf{Route Completion (RC) of 89.04\%} on the \textit{CARLA Town05 Short Benchmark}, FASIONAD demonstrates superior performance across standard benchmarks. Its dual-pathway design enhances adaptability and supports integration with multi-sensor navigation systems, positioning it for real-time deployment in both single-agent and fleet-based scenarios. Future work will focus on expanding FASIONAD’s robustness in rural and unstructured environments and integrating additional sensor modalities such as \textbf{LiDAR and radar}. Furthermore, ongoing research into \textbf{reinforcement learning from human feedback (RLHF)} and fleet coordination will explore avenues for optimizing performance in complex, multi-agent contexts. \textbf{FASIONAD} is poised to meet the operational demands of diverse, real-world driving environments, representing a significant advancement in autonomous navigation.

{
    \small
    \bibliographystyle{ieeenat_fullname}
    \bibliography{main}
}

\appendix
\section{More Details about Model Design}
\begin{figure*}[htbp]
    \centering
    \vspace{-10pt}
    \includegraphics[width=\linewidth]{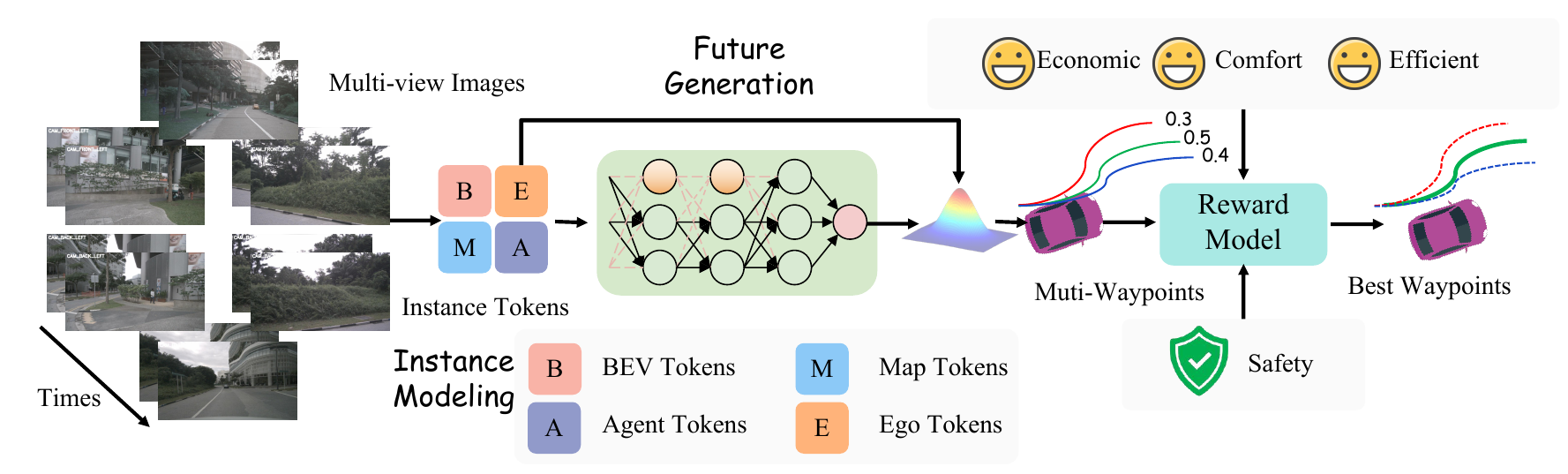}
    \vspace{-20pt}
    \caption{FASIONAD's fast pathway. We adopt the BEVFormer approach \cite{li2022bevformer} to extract BEV features, denoted as $\mathbf{B} \in \mathbb{R}^{bs \times H \times W \times C}$, which encapsulate the environmental topology. Self-attention and cross-attention mechanisms are used to model the interactions between different tokens (agent tokens $\textbf{T}_{\textbf{A}}$, map tokens $\textbf{T}_{\textbf{M}}$, and ego tokens $\textbf{T}_{\textbf{E}}$). The generative framework, similar to that in GenAD \cite{GenAD}, is employed in the trajectory generator. Finally, the reward model provides a reward for each trajectory and selects the best trajectory as the fast pathway output.}
    \vspace{-5pt}
    \label{fig:bev1}
\end{figure*}
\subsection{Instance Modeling}
The first step in the Fast Pathway is to process sensor inputs to obtain high-level descriptions of the surrounding environment. Inspired by the decision-making processes of human drivers, we categorize the information needed for decision-making into two levels: low-level perception information (\textit{what do we observe?}) and high-level perception information (\textit{understanding the interactions among observed elements}). Low-level perception information includes details about traffic participants and map features, while high-level perception information captures the interactions between these elements, as shown in Fig.\ref{fig:bev1}.

\textbf{BEV Encoder.} Following this framework, we adopt a vision-centric perception pipeline, initially extracting Bird’s Eye View (BEV) features, which serve as a foundation for refining both low-level and high-level perception information. We adopt the BEVFormer approach \cite{li2022bevformer} to extract BEV features, denoted as $\mathbf{B} \in \mathbb{R}^{bs\times H \times W \times C}$( where \( bs \) represents the batch size and \( C \) the feature dimension), which encapsulate the environmental topology. This process can be formulated as:
\begin{equation}
    \mathcal{F}(I_t)=\mathbf{B}_t
    \end{equation}
\paragraph{Instance-centric Token Modeling.}
Since semantic map elements are typically sparse in the BEV space, we represent low-level perception information using sparse map tokens and agent tokens. These tokens are updated through attention mechanisms with BEV tokens, following the approach in \cite{jiang2023vad}. Specifically, the updated Agent Tokens, now incorporating motion information, are concatenated with the Ego Token. Self-attention is then applied to capture high-order interactions among different traffic participants. Additionally, Map Tokens engage in cross-attention with tokens that encapsulate interactions between traffic participants. This final cross-attention step enriches the tokens with semantic map information, resulting in an instance-centric scene representation that captures both agent interactions and contextual map features. This representation provides a comprehensive and compact description of the autonomous driving scenario. The modeling process is represented as follows:
\begin{equation}
    \textbf{T}_i = \text{CA}(T_i, B),\quad i \in \{\mathbf{A}, \mathbf{M}\} 
\end{equation}
\begin{equation}
      \textbf{T}_{\mathcal{\mathbf{I}}} = \text{CA}(\text{SA}([\textbf{T}_\mathbf{A}', \textbf{T}_\mathbf{E}]), \textbf{T}_\textbf{M}')
\end{equation}
where \( T_{\mathcal{\mathbf{I}}}, T_{\mathbf{A}}, T_{\mathbf{M}} \) denote instance tokens, agent tokens, and map tokens, respectively. CA denotes the cross attention mechanism.

\paragraph{Cross-Attention Mechanism.} 
To incorporate contextual information, cross-attention is applied between the BEV tokens \( \mathbf{T}_{\text{bev}} \) and feature sets \( \mathbf{F}_a \), \( \mathbf{F}_e \), and \( \mathbf{F}_m \), which represent agents, the ego vehicle, and map features, respectively:
\begin{align}
    \mathbf{T}_{\textbf{A}} &= \text{softmax} \left( \frac{\mathbf{Q} \mathbf{K}_{\textbf{A}}^\top}{\sqrt{d}} \right) \mathbf{V}_{\textbf{A}}, \\
    \mathbf{T}_{\textbf{E}} &= \text{softmax} \left( \frac{\mathbf{Q} \mathbf{K}_e^\top}{\sqrt{d}} \right) \mathbf{V}_{\textbf{E}}, \\
    \mathbf{T}_{\textbf{M}} &= \text{softmax} \left( \frac{\mathbf{Q} \mathbf{K}_{\textbf{M}}^\top}{\sqrt{d}} \right) \mathbf{V}_{\textbf{M}},
\end{align}
where \( \mathbf{Q} = \mathbf{W}_q \mathbf{T}_{\text{bev}} \), \( \mathbf{K}_* = \mathbf{W}_k \mathbf{F}_* \), and \( \mathbf{V}_* = \mathbf{W}_v \mathbf{F}_* \) for \( * \in \{{\textbf{A}}, {\textbf{E}}, {\textbf{M}}\} \). Here, \( \mathbf{W}_q \), \( \mathbf{W}_k \), and \( \mathbf{W}_v \) are learnable projection matrices, and \( d \) is a scaling factor.

\paragraph{Trajectory Generator.}
The trajectory generator outputs waypoint predictions \( \mathcal{W} = \{\mathbf{w}_t\}_{t=1}^{T} \), with each waypoint \( \mathbf{w}_t = (x_t, y_t) \) representing a spatial position in BEV coordinates. To capture interactions among traffic participants, we adopt a generative framework inspired by GenAD \cite{GenAD}, modeling trajectory prediction as a future trajectory generation problem.
\begin{equation}
    p(\mathcal{Z}|\mathcal{I}(\mathcal{B'})) \sim \mathcal{N}(\mu_f, \sigma_f^2)
\end{equation}
where \( \mathcal{N}(\mu, \sigma^2) \) represents a Gaussian distribution parameterized by mean \( \mu \) and variance \( \sigma^2 \). This learned distribution \( p(\mathcal{Z}|\mathcal{I}(\mathcal{B'})) \) captures prior knowledge of the ground-truth motion field, enhancing the realism of motion predictions.
\begin{equation}
    p(\mathcal{W}_{T+1}|\mathcal{W}_T, \ldots, \mathcal{W}_{T+f-1}, \mathcal{Z})
\end{equation}

\paragraph{Reward Model.}
\begin{align}
    \mathcal{F}_{\text{Reward}} = &\ \alpha_{\text{safety}} C_{\text{safety}} 
    + \alpha_{\text{comfort}} C_{\text{comfort}} \notag \\
    &+ \alpha_{\text{efficiency}} C_{\text{efficiency}} 
    + \alpha_{\text{economic}} C_{\text{economic}}
\end{align}
where \( \alpha_{\text{safety}}, \alpha_{\text{comfort}}, \alpha_{\text{efficiency}}, \alpha_{\text{economic}} \) are weighting coefficients for each aspect.

\textbf{Safety Component.} \( C_{\text{safety}} \) considers collision risk, distance to obstacles, deviation from the desired path, and speed constraints:
\begin{align}
    C_{\text{safety}} = &\ W_{\text{coll}} C_{\text{coll}} 
    + W_{\text{dist}} C_{\text{dist}} \notag \\
    &+ W_{\text{deviation}} C_{\text{deviation}} 
    + W_{\text{speed}} C_{\text{speed}}.
\end{align}
where \( C_{\text{coll}} = e^{\left(-\frac{d_{\text{coll}}}{\sigma_{\text{coll}}}\right)} \) measures the exponential decay of collision risk with distance \( d_{\text{coll}} \), and \( C_{\text{deviation}} = \sum_{i=1}^{N} \left(1 - \cos(\theta_i - \theta_{\text{target}})\right) \) evaluates directional alignment with the desired path. The terms \( C_{\text{dist}} \) and \( C_{\text{speed}} \) penalize the deviation from ideal positioning and target speed, respectively, using VAD/GenAD for critical areas.

\textbf{Comfort Component.} \( C_{\text{comfort}} \) evaluates smoothness and comfort in terms of lateral and longitudinal accelerations:
\begin{equation}
    C_{\text{comfort}} = W_{\text{lat}} C_{\text{lat}} + W_{\text{lon}} C_{\text{lon}} + W_{\text{cent}} C_{\text{cent}},
\end{equation}
where \( C_{\text{lat}} \) and \( C_{\text{lon}} \) are the lateral and longitudinal acceleration terms, and \( C_{\text{cent}} \) accounts for centrifugal forces, promoting comfortable maneuvers.

\textbf{Efficiency Component.} \( C_{\text{efficiency}} \) focuses on maintaining optimal speed and minimizing travel time:
\begin{equation}
    C_{\text{efficiency}} = W_{\text{speed}} C_{\text{speed}} + W_{\text{time}} C_{\text{time}},
\end{equation}
where \( C_{\text{speed}} \) penalizes deviations from the target speed \( V_{\text{target}} \), and \( C_{\text{time}} \) incentivizes efficient travel times.

\textbf{Economic Component.} \( C_{\text{economic}} \) reduces fuel consumption or energy usage:
\begin{equation}
    C_{\text{economic}} = \begin{cases} 
       k_v \cdot V + c_v, & \text{for speed} \\
       k_a \cdot A + c_a, & \text{for acceleration}
    \end{cases}
\end{equation}
where \( k_v \) and \( k_a \) are coefficients for speed and acceleration, with constants \( c_v \) and \( c_a \) representing baseline energy consumption.

Once the rewards \( r_i = C_{\text{total}} \) are computed for each candidate trajectory, the model selects the \textit{top-K} trajectories based on the highest reward scores. If the rewards are uniformly low or the distribution indicates high uncertainty, a \textit{reward distribution check} is performed to determine if VLMs are needed for further refinement. The VLM generates a Visual Prompt to assess collision risks, traffic rule violations, and environmental context, updating the semantic embedding \( \psi \) based on this feedback. For example, the VLM may evaluate if the vehicle is positioned correctly to avoid violations like crossing road lines or exceeding speed limits, especially when overtaking or approaching intersections. The model then adjusts \( V_{\text{target}} \) based on these checks, incorporating surrounding vehicle speed and contextual speed limits.

\subsection{VLMs QA}

The Slow Pathway in \textbf{FASIONAD} leverages Vision-Language Models (VLMs) to perform advanced reasoning in complex scenarios. This section details its inputs, the modeling process to derive intermediate tokens, and demonstrates its functionality with a specific example derived from the main paper and JSON data.

\subsubsection{Inputs}

The Slow Pathway processes raw visual inputs \(I_t\) and generates intermediate representations. It also relies on spatial and semantic prompts derived from Fast Pathway outputs. Key inputs include:

\begin{itemize}
    \item \textbf{Multi-view Images} (\(I_t\)): 
    Front-view sensory data collected from cameras at time \(t\), providing raw visual information for scene understanding.
    
    \item \textbf{BEV and Visual Prompts}:
    Derived from the Fast Pathway's trajectory outputs and 3D bounding boxes (\(bbox_{3D}\)). These prompts provide spatial and dynamic context:
    \begin{itemize}
        \item BEV Prompt: Encodes the spatial relationships and interactions between agents, such as vehicles and pedestrians.
        \item Visual Prompt: Highlights projected trajectories and potential risks, overlaying them onto the camera view for enhanced interpretability.
    \end{itemize}
\end{itemize}

\subsubsection{Modeled Representations}

Intermediate tokens and semantic features are generated during the modeling process to support decision-making:

\begin{itemize}
    \item \textbf{Instance Tokens}: Derived via self-attention and cross-attention mechanisms. These include:
    \begin{itemize}
        \item \textbf{Ego Tokens} (\(E\)): Encodes the ego vehicle's state, including position, velocity, and orientation.
        \item \textbf{Agent Tokens} (\(A\)): Represent dynamic agents and their predicted trajectories.
        \item \textbf{Map Tokens} (\(M\)): Capture static environmental features like traffic signals, lanes, and intersections.
    \end{itemize}
    
    \item \textbf{Planning State Vectors} (\(\mathcal{P}_t\)): \(K\)-dimensional binary vectors representing environmental conditions relevant to planning, e.g., pedestrian presence (\(1\)), no obstacles (\(0\)), red light (\(1\)).
    
    \item \textbf{Meta-Actions} (\(\mathcal{A}_t\)): High-level planning actions, such as lane changes or stopping, extracted via structured reasoning over instance and prompt data.
\end{itemize}

\subsubsection{Decision Process}

The Slow Pathway interprets environmental context to refine trajectory predictions and meta-actions. This includes:

\begin{itemize}
    \item \textbf{Uncertainty Estimation}: A threshold-based mechanism evaluates the uncertainty in waypoint predictions from the Fast Pathway. If uncertainty exceeds the threshold, the Slow Pathway is activated for detailed analysis.
    \item \textbf{QA Reasoning}: Predefined Question-Answering tasks, such as "What traffic rules apply?" and "Are there potential conflicts?", guide the generation of structured planning outputs.
\end{itemize}

\subsubsection{Outputs}

\begin{itemize}
    \item \textbf{Planning State} (\(\mathcal{P}_t\)): Binary vectors encoding key scene-specific conditions.
    \item \textbf{Meta-Actions} (\(\mathcal{A}_t\)): High-level decisions directly guiding trajectory refinement.
    \item \textbf{Refined Trajectories} (\(\Delta \mathcal{W}\)): Adjusted waypoints integrating feedback from VLMs.
    \item \textbf{Semantic Feedback}: Descriptive language-based recommendations for future actions, e.g., "Yield to pedestrian, then prepare to stop."
\end{itemize}

\subsubsection{Example: Complex Intersection Navigation}

\paragraph{Scenario:} The vehicle approaches an intersection with pedestrians and other vehicles under rainy conditions.

\paragraph{Inputs:}
\begin{itemize}
    \item \textbf{Multi-view images} (\(I_t\)): Capturing reduced visibility, crosswalk activity, and adjacent vehicle signaling.
    \item \textbf{BEV Tokens} (\(B\)): Encodes pedestrian positions and traffic light states.
    \item \textbf{Prompts}: Emphasizing risks (collision or rule violation) and spatial relationships.
\end{itemize}

\paragraph{Outputs:}
\begin{itemize}
    \item \textbf{Planning State}:
    \begin{equation}
        \mathcal{P}_t = [1 \ (\text{pedestrian}), \ 1 \ (\text{adjacent vehicle}), \ 1 \ (\text{red light})]
    \end{equation}
    \item \textbf{Meta-Actions}:
    \begin{equation}
        \mathcal{A}_t = [\text{Stop}, \ \text{Wait}, \ \text{Prepare\_Turn}]
    \end{equation}
    \item \textbf{Refined Trajectory}:
    \begin{equation}
        \mathcal{W}_t = \{(x_{t+1}, y_{t+1}), \dots, (x_{t+n}, y_{t+n})\}
    \end{equation}
    \item \textbf{Semantic Feedback}:
    \begin{quote}
        \textit{"Stop at crosswalk. Wait for pedestrians to clear. Yield to adjacent vehicle turning left."}
    \end{quote}
\end{itemize}

\subsection{Slow Pathway Feedback}
\paragraph{Information Bottleneck.}To reduce noise and enhance feature alignment, we incorporate an information bottleneck objective, which seeks to learn a representation that minimizes the correlation between the representation and the input while maximizing the correlation between the representation and the class.This is achieved through the following optimization:

\begin{equation}
\max \; I(y_t, y_i) - \beta \, I(y_i, H),
\end{equation}
where \( I(y_t, y_i) \) measures the relevant mutual information between \( y_t \) and \( y_i \), ensuring that the representation retains useful class-specific information, while \( I(y_i, H) \) captures the mutual information between \( y_i \) and extraneous variables \( H \), which is minimized to avoid noise. The information bottleneck principle ensures that \( y_i \) serves as a compact yet effective representation that focuses on class-relevant features, enhancing the alignment with \( y_t \).

In the information bottleneck, we configure both the encoder and decoder of the information bottleneck by using a 3-layer MLPs with a hidden size of 512 and LeakyReLU\cite{leakyrelu} as the activation function. 
\paragraph{High-level plans. }
To update the Ego tokens \(\textbf{T}_{\textbf{E}}\), multi-head attention(MHA) is used, with the Ego tokens directly incorporating distilled semantic information from the high-level actions embedding with CLIP encoder. The updates are defined as follows:
\begin{equation}
    \textbf{T}'_{\textbf{E}} = \text{MHA}(\textbf{T}_{\textbf{E}}, \mathcal{A}^{\text{distill}}) + \textbf{T}_{\textbf{E}}
\end{equation}

where \( T_{\text{sem}} \) is the token refined by the semantic embedding \( \psi \), and \( y_t^{\text{distill}} \) is the distilled semantic vector generated by the model \( f_{\text{distill}} \) as an approximation of \( y_t \). By directly incorporating \( y_t^{\text{distill}} \) into the update of the Ego tokens, we ensure that the Ego tokens remain aligned with the latest semantic context and driving intent as interpreted by the VLM, while reducing computational load in subsequent steps

\begin{figure*}[htbp]
    \centering
    \vspace{-10pt}
    \includegraphics[width=\linewidth]{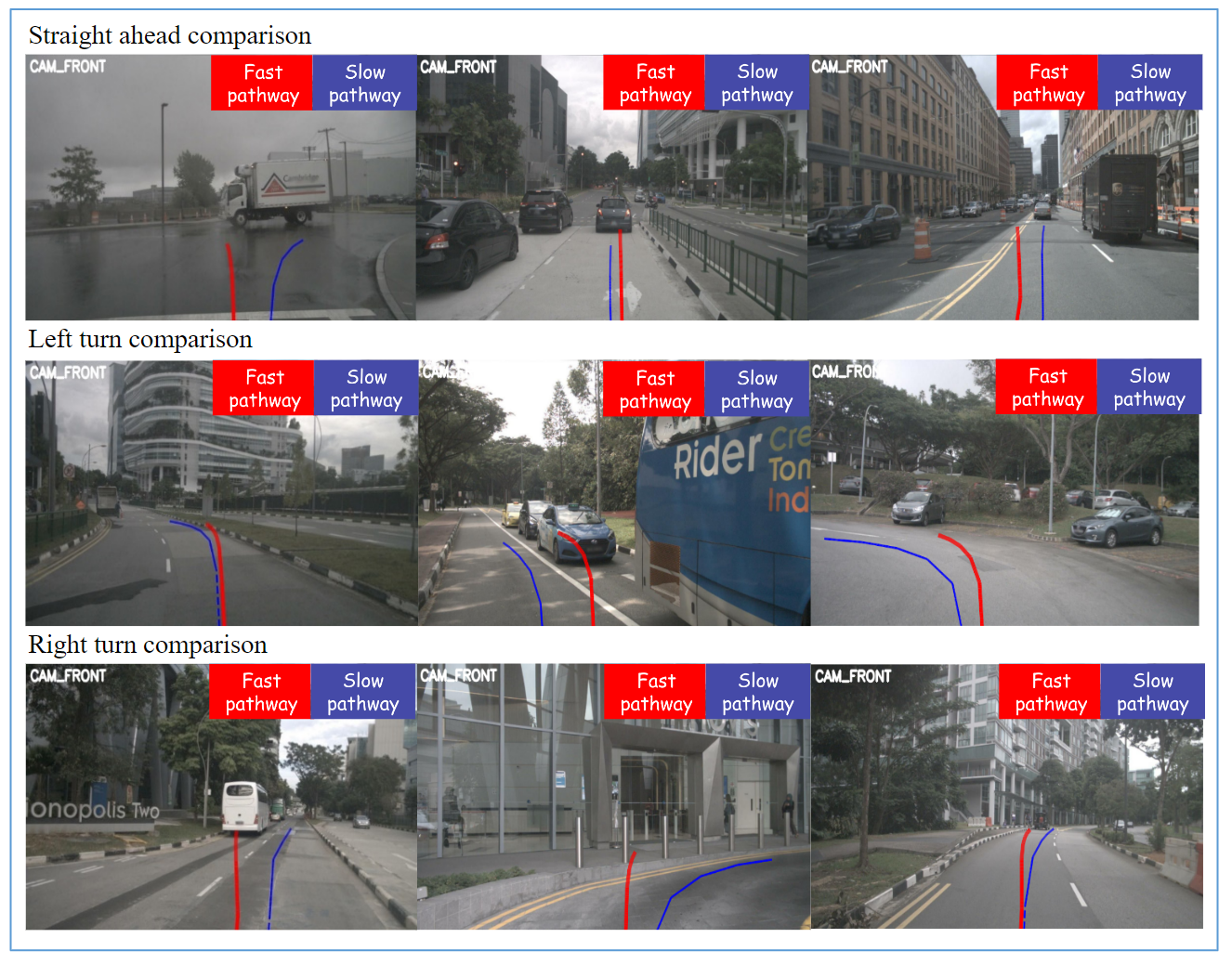}
    \vspace{-20pt}
    \caption{This figure illustrates FASIONAD's adaptive feedback framework across various driving environments, showcasing comparisons between the "Fast pathway" and "Slow pathway" in three categories: straight driving, left turns, and right turns. Each category includes three representative examples. The straight driving scenarios highlight the system's ability to handle varying traffic density, road geometry, and dynamic obstacles, where the "Fast pathway" focuses on immediate trajectory optimization, and the "Slow pathway" integrates longer-term considerations for smoother and safer navigation. The left turn scenarios emphasize responses to complex intersections, occluded views, and interactions with other road users, with the "Fast pathway" prioritizing efficiency and the "Slow pathway" adapting for increased safety and compliance. The right turn scenarios demonstrate FASIONAD's capability to navigate challenging environments, such as narrow lanes, pedestrian zones, and tight corners, where the "Slow pathway" ensures precise control and collision avoidance.}
    \vspace{-5pt}
    \label{fig:bev}
\end{figure*}

\subsection{Laplace-Based Uncertainty Estimation}

We propose the probabilistic modeling of the reward based on the Laplace distribution for uncertainty estimation. The probability density function (PDF) of a univariate Laplace distribution is:

\begin{equation}
    f(x \mid \mu, b) = \frac{1}{2b} \exp\left(-\frac{|x - \mu|}{b}\right)
\end{equation}

where \(x\) is the random variable, \(\mu\) is the location parameter, and \(b > 0\) is the scale parameter. This describes the distribution of a single variable.

Here, the target is the reward function \(\mathbf{r}_t\) at each time step. Assuming that the reward \(\mathbf{r}_t\) is a multivariate vector and the dimensions of \(\mathbf{r}_t\) are independent, each dimension follows a univariate Laplace distribution. The joint probability density for a single time step \(t\) can be expressed as:

\begin{equation}
    p(\mathbf{r}_t \mid \hat{\mathbf{r}}_t, b) = \prod_{j=1}^{d} \frac{1}{2b} \exp\left(-\frac{|r_{t,j} - \hat{r}_{t,j}|}{b}\right)
\end{equation}

where $d$ is the dimension of the reward vector \(\mathbf{r}_t\); $r_{t,j}$ is the \(j\)-th component of the reward vector at time \(t\); $\hat{r}_{t,j}$ is the \(j\)-th component of the predicted reward at time \(t\), \(b > 0\) is the scale parameter, controlling the width of the distribution.

If \(d = 1\), i.e., the reward is scalar, the formula simplifies to:

\begin{equation}
    p(\mathbf{r}_t \mid \hat{\mathbf{r}}_t, b) = \frac{1}{2b} \exp\left(-\frac{\|\mathbf{r}_t - \hat{\mathbf{r}}_t\|_1}{b}\right)
\end{equation}

where \(\|\mathbf{r}_t - \hat{\mathbf{r}}_t\|_1\) is the \(L_1\)-norm (absolute deviation) of the reward.

Extending this to the entire time sequence, we assume that the rewards \(\mathbf{r}_t\) at each time step are independent. The joint probability density over the sequence is:

\begin{equation}
p(\text{Reward} \mid \Theta) = \prod_{t=1}^{T} p(\mathbf{r}_t \mid \hat{\mathbf{r}}_t, b)
\end{equation}

where \(\Theta\) represents the model parameters that define the predicted rewards \(\hat{\mathbf{r}}_t\). Substituting the single-step probability density yields:

\begin{equation}
    p(\text{Reward} \mid \Theta) = \prod_{t=1}^{T} \frac{1}{2b} \exp\left(-\frac{\|\mathbf{r}_t - \hat{\mathbf{r}}_t\|_1}{b}\right)
\end{equation}

The Laplace distribution is chosen here because of its heavy-tailed nature and robustness to outliers. In dynamic driving scenarios, reward signals may be affected by noise or anomalies. Compared to the squared deviation (\(L_2\)-norm) in Gaussian distributions, the absolute deviation (\(L_1\)-norm) used in the Laplace distribution is more robust to outliers, avoiding excessive amplification of large deviations. Additionally, the Laplace distribution is parameterized by a single scale parameter \(b\), making it simpler to train and optimize. This makes the model suitable for handling dynamic and uncertain scenarios.

\section{More Details about Loss Function}

\textbf{Maximum Likelihood Estimation (MLE) Loss.} In the context of language models, the MLE loss is typically used to optimize the likelihood of the target sequence given the input sequence. The MLE loss for a token sequence $\mathbf{s} = \{s_1, s_2, ..., s_T\}$ is defined as:
\begin{equation}
    \mathcal{L}_{\text{MLE}} = -\sum_{t=1}^{T} \log P(s_t | s_1, s_2, ..., s_{t-1})
\end{equation}
where $P(s_t | s_1, s_2, ..., s_{t-1})$ is the conditional probability of the token $s_t$ given the previous tokens in the sequence. This loss is minimized during training to maximize the likelihood of generating the correct sequence, making it suitable for auto-regressive learning in tasks like trajectory prediction.

\section{More Details about Training Strategy.}

Our training process is divided into three stages: (1) training the Fast Pathway for generating reasonable trajectories and a robust reward function, (2) fine-tuning Vision-Language Models (VLMs) to output structured vector representations, and (3) joint training of the Fast and Slow Pathways to align feedback and improve performance in complex scenarios.

\subsection{Stage 1: Training the Fast Pathway}

The first stage focuses on learning robust trajectory generation and designing a reward model that evaluates safety, efficiency, and comfort.

\textbf{Objective:} \begin{itemize} \item Train the Fast Pathway to output accurate trajectories $\mathcal{W}$ for real-time decision-making. \item Optimize a reward model $\mathcal{F}_{\text{Reward}}$ to score trajectories based on multiple factors. \end{itemize}

The Fast Pathway processes raw sensory data and generates intermediate representations for trajectory prediction. The key components are described below:

\begin{itemize}
    \item \textbf{Inputs:}
    \begin{itemize}
        \item \textbf{Multi-view images} (\(I_t\)): The raw sensory input from front-view cameras at time \(t\), providing visual information for scene understanding and real-time decision-making.
    \end{itemize}

    \item \textbf{Modeled Representations (Generated by the Fast Pathway):}
    \begin{itemize}
        \item \textbf{BEV tokens} (\(\mathbf{B}\)): Encoded spatial representations of the scene in a Bird's Eye View format, capturing the positions and movements of surrounding entities. These are derived from BEV encoding layers.
        \item \textbf{Instance tokens} (\(\mathbf{T}_{\mathcal{I}}\)): High-level features derived from \(\mathbf{B}\) through modeling steps that involve self-attention and cross-attention mechanisms. These tokens include:
        \begin{itemize}
            \item \textbf{Ego Tokens} (\(\mathbf{E}\)): Represent the ego vehicle's state, such as position, velocity, and orientation.
            \item \textbf{Agent Tokens} (\(\mathbf{A}\)): Encode information about dynamic agents, including nearby vehicles and pedestrians.
            \item \textbf{Map Tokens} (\(\mathbf{M}\)): Describe static environmental features such as lanes, traffic signals, and intersections.
        \end{itemize}
    \end{itemize}
\end{itemize}
\textbf{Outputs:}  
\begin{itemize}  
    \item Predicted waypoints \(\mathcal{W}\): A sequence of waypoints representing the planned trajectory.  
    \item Reward scores \(r_t\): Evaluations of safety, efficiency, and comfort for the predicted trajectories.  
\end{itemize}  

\textbf{Loss Function:}  
\begin{equation}  
\mathcal{L}_{\text{fast}} = \lambda_{\text{plan}} \mathcal{L}_{\text{plan}} + \lambda_{\text{det}} \mathcal{L}_{\text{det}} + \lambda_{\text{seg}} \mathcal{L}_{\text{seg}} + \lambda_{\text{reward}} \mathcal{L}_{\text{reward}},  
\end{equation}  
where:  
\begin{itemize}  
    \item \(\mathcal{L}_{\text{plan}}\): Trajectory prediction loss, computed as L2 distance between predicted and ground-truth waypoints.  
    \item \(\mathcal{L}_{\text{det}}\): 3D detection loss.  
    \item \(\mathcal{L}_{\text{seg}}\): Map segmentation loss.  
    \item \(\mathcal{L}_{\text{reward}}\): Reward loss, defined as:  
    \begin{equation}  
    \mathcal{L}_{\text{reward}} = \left| \mathcal{F}_{\text{Reward}}^{\text{pred}} - \mathcal{F}_{\text{Reward}}^{\text{gt}} \right|_1,  
    \end{equation}  
    where \(\mathcal{F}_{\text{Reward}}^{\text{gt}}\) is the ground truth reward, calculated based on safety, efficiency, and comfort components, and \(\mathcal{F}_{\text{Reward}}^{\text{pred}}\) is the predicted reward score.  
\end{itemize}  

\textbf{Reward Model Training:}  
The reward function \(\mathcal{F}_{\text{Reward}}\) evaluates trajectory quality:  
\begin{equation}  
\mathcal{F}_{\text{Reward}} = \alpha_{\text{safety}} C_{\text{safety}} + \alpha_{\text{comfort}} C_{\text{comfort}} + \alpha_{\text{efficiency}} C_{\text{efficiency}},  
\end{equation}  
where:  
\begin{itemize}  
    \item \(C_{\text{safety}}\): Penalizes unsafe actions, such as collisions or near misses.  
    \item \(C_{\text{comfort}}\): Encourages smooth trajectories by minimizing abrupt accelerations or turns.  
    \item \(C_{\text{efficiency}}\): Rewards timely and energy-efficient travel.  
    \item \(\alpha_{\text{safety}}, \alpha_{\text{comfort}}, \alpha_{\text{efficiency}}\): Weights for each component.  
\end{itemize}  

\textbf{Handling Uncertainty in Feedback:}  
During training, feedback from the VLMs is only updated once per sample. If the predicted reward scores exhibit significant uncertainty or unreliability (e.g., hallucinations in the feedback causing poor trajectory alignment), the system discards the feedback and does not update the trajectory for that sample.

\subsection{Stage 2: Fine-Tuning VLMs}
The second stage focuses on fine-tuning Vision-Language Models (VLMs) to generate structured vector representations, enhancing the Slow Pathway's ability to provide high-quality feedback for decision-making.

\textbf{Objective:}  
The goal of this stage is to train VLMs to interpret visual and BEV prompts, generating structured feedback vectors (\( \psi \)) and high-level plans (\( \mathcal{A} \)).

\textbf{Inputs:}  
The fine-tuning process utilizes the following inputs:  
\begin{itemize}
    \item Visual prompts (\( P_{\text{visual}} \)): Front-view images containing trajectory information.
    \item BEV prompts (\( P_{\text{BEV}} \)): Bird’s Eye View representations for spatial context.
    \item Question-answering tasks: Designed to simulate human-like reasoning and interaction with driving scenarios.
\end{itemize}

\textbf{Outputs:}  
The outputs of this stage include:  
\begin{itemize}
    \item Planning states (\( \mathcal{P}_t \)): Representations of environmental context relevant to decision-making.
    \item Meta-actions (\( \mathcal{A}_t \)): High-level actions derived from structured semantic understanding.
\end{itemize}

\textbf{Loss Function:}  
The fine-tuning process is guided by the following loss function:
\begin{equation}
\mathcal{L}_{\text{slow}} = \lambda_{\text{MLE}} \mathcal{L}_{\text{MLE}} + \lambda_{\text{rvlm}} \mathcal{L}_{\text{rvlm}},
\end{equation}
where:
\begin{itemize}
    \item \( \mathcal{L}_{\text{MLE}} = -\sum_{t} \log P(s_t | s_{<t}) \): Maximum Likelihood Estimation loss for token prediction, ensuring the model learns to generate accurate sequential representations.
    \item \( \mathcal{L}_{\text{rvlm}} = \text{Reward}(\mathbf{s}^{1:T}) \cdot \Phi(\mathbf{s}^{T} | \mathbf{s}^{<T}) \): Reward-guided loss that aligns outputs with trajectory-based supervision, enhancing consistency and alignment with desired planning goals.
\end{itemize}

By fine-tuning the VLMs with these inputs, outputs, and loss functions, the model achieves a robust capability to interpret diverse driving scenarios and provide actionable, structured feedback to the Fast Pathway.

\subsection{Stage 3: Joint Training of Fast and Slow Pathways}

The final stage focuses on integrating the Slow Pathway's reasoning-based feedback into the Fast Pathway's real-time trajectory generation. This process ensures that the system combines the efficiency of the Fast Pathway with the contextual reasoning and adaptability of the Slow Pathway, aligning their outputs for improved overall performance.

\textbf{Objective:} \begin{itemize} \item Refine the Fast Pathway's real-time trajectory generation by incorporating structured feedback from the Slow Pathway. \item Achieve alignment between the planning states $\mathcal{P}_t$, meta-actions $\mathcal{A}_t$, and waypoints $\mathcal{W}$ generated by both pathways. \item Minimize inconsistencies between Fast Pathway outputs and the Slow Pathway's high-level plans, especially in complex or safety-critical scenarios. \end{itemize}

\textbf{Alignment Definition:} Alignment in this context refers to ensuring that the outputs of the Fast Pathway (e.g., planning states $\mathcal{P}_t$ and meta-actions $\mathcal{A}_t$) are consistent with the reasoning-based decisions generated by the Slow Pathway ($\hat{\mathcal{P}}_t$ and $\hat{\mathcal{A}}_t$). This alignment addresses the following: \begin{itemize} \item Planning States Alignment ($\mathcal{P}_t$ vs. $\hat{\mathcal{P}}_t$): Ensures that the Fast Pathway accurately interprets and integrates the situational understanding provided by the Slow Pathway, such as obstacles, dynamic agent behavior, and environmental factors. \item Meta-Actions Alignment ($\mathcal{A}_t$ vs. $\hat{\mathcal{A}}_t$): Guarantees that the high-level decisions (e.g., "stop at intersection," "overtake") recommended by the Slow Pathway are respected and reflected in the Fast Pathway's trajectory outputs. \item Trajectory Refinement: Enhances the trajectory $\mathcal{W}$ generated by the Fast Pathway by incorporating corrections or modifications suggested by the Slow Pathway, such as reducing collision risks or ensuring regulatory compliance. \end{itemize}

\textbf{Inputs:} \begin{itemize} \item Waypoints from the Fast Pathway ($\mathcal{W}$): The initial trajectory predictions generated based on real-time inputs. \item Feedback Vectors from the Slow Pathway: Includes refined planning states $\hat{\mathcal{P}}_t$ and high-level meta-actions $\hat{\mathcal{A}}_t$, derived from VLM-based reasoning. \end{itemize}

\textbf{Outputs:} \begin{itemize} \item Updated Trajectories ($\mathcal{W'}$): Trajectories that incorporate feedback to enhance safety, efficiency, and adaptability. \item Refined Reward Scores ($r_t$): Scores that reflect the quality of the updated trajectories based on multiple factors, including safety and compliance. \end{itemize}

\textbf{Loss Function:} The joint training process optimizes a combined loss function: \begin{equation} \mathcal{L}_{\text{joint}} = \mathcal{L}_{\text{fast}} + \mathcal{L}_{\text{slow}} + \lambda_{\text{align}} \mathcal{L}_{\text{align}}, \end{equation} where: \begin{itemize} \item $\mathcal{L}_{\text{fast}}$: Loss associated with the Fast Pathway, including trajectory prediction, detection, and segmentation errors. \item $\mathcal{L}_{\text{slow}}$: Loss associated with the Slow Pathway, including MLE-based and reward-guided loss components. \item $\mathcal{L}_{\text{align}}$: Ensures alignment between the two pathways, defined as: \begin{equation} \mathcal{L}_{\text{align}} = | \mathcal{P}_t - \hat{\mathcal{P}}_t |_2^2 + | \mathcal{A}_t - \hat{\mathcal{A}}_t |_2^2. \end{equation} This term penalizes discrepancies between the planning states and meta-actions of the two pathways. \end{itemize}

\textbf{Detailed Process:}

Trajectory Alignment:

For each predicted trajectory $\mathcal{W}$ from the Fast Pathway, feedback from the Slow Pathway adjusts the waypoints based on contextual reasoning.
For example, if the Fast Pathway suggests an overly aggressive lane change, the Slow Pathway may recommend a more conservative maneuver to ensure safety.
Iterative Feedback Refinement:

The Slow Pathway’s feedback is incorporated iteratively, with adjustments made to both the planning states and meta-actions of the Fast Pathway.
This process ensures that the refined trajectories respect the broader contextual understanding provided by the Slow Pathway.
Safety and Performance Optimization:

By integrating Slow Pathway insights, the system reduces risks associated with long-tail events (e.g., unexpected pedestrian crossings) and improves compliance with traffic regulations.

\paragraph{Example Scenario: Emergency Braking at Intersection.} Consider a scenario where the Fast Pathway predicts a trajectory to proceed through an intersection without noticing a stop sign partially obscured by a tree. The joint training ensures the following: \begin{enumerate} \item Fast Pathway Prediction: Generates a trajectory $\mathcal{W}$ that continues through the intersection without stopping. \item Slow Pathway Feedback: \begin{itemize} \item Planning state $\hat{\mathcal{P}}_t$: Indicates the presence of a stop sign and high traffic density at the intersection. \item Meta-action $\hat{\mathcal{A}}_t$: Recommends stopping at the intersection and waiting for clearance. \end{itemize} \item Trajectory Refinement: The Fast Pathway adjusts its trajectory $\mathcal{W'}$ to include a stop action at the intersection, ensuring compliance and safety. \item Outcome: The refined trajectory avoids a potential collision or traffic violation, demonstrating the benefits of alignment. \end{enumerate}

\section{More Experimental Setting Details}
\paragraph{Fast pathway implementation.}We adopted ResNet50\cite{he2016deep} as the backbone network to extract image features. We take as input images with a resolution of 640 × 360 and use a 200 × 200 BEV representation
to perceive the surrounding scene. For fair comparisons, we
basically use the same hyperparameters as VAD-tiny\cite{jiang2023vad}.
We ffxed the number of BEV tokens, map tokens, and agent
tokens to 100 × 100, 100, and 300, respectively. Each map
token contains 20 point tokens to represent a map point in
the BEV space. We set the hidden dimension of each BEV,
point, agent, ego, and instance token to 256. We set the \(\alpha_{\text{safety}}=2\),\(\alpha_{\text{comfort}}=\alpha_{\text{efficiency}}=\alpha_{\text{economic}}=1\) in reward function. 

For training, we set the loss balance factors to 1 and
use the AdamW\cite{AdamW2019} optimizer with a cosine learning rate
scheduler\cite{loshchilov2016sgdr}. We set the initial learning rate to 2 × 10-4 and a weight decay of 0.01. By default, we trained our
FASIONAD for 30 epochs with 8 NVIDIA Tesla A100 GPUs
and adopted a total batch size of 8.

\end{document}